\documentclass{article}

\usepackage[final]{corl_2025} %
\usepackage{graphicx}
\usepackage{booktabs}
\usepackage{pifont}
\usepackage{wrapfig}
\usepackage{subcaption}
\usepackage{amsmath}
\usepackage{amssymb}
\usepackage{placeins}
\usepackage[T1]{fontenc}
\usepackage{cleveref}
\usepackage{array}
\usepackage{tabularx}
\usepackage{tikz}
\newcolumntype{Y}{>{\centering\arraybackslash}X} 
\crefname{equation}{Eq.}{Eqs.}
\Crefname{equation}{Eq.}{Eqs.}
\crefname{figure}{Fig.}{Figs.}
\Crefname{figure}{Fig.}{Figs.}

\DeclareUrlCommand\url{\color{ellisred}}
\hypersetup{
  colorlinks=true,
  urlcolor=ellisred, 
  linkcolor=ellisred,
  citecolor=ellisblue,
  pdfborder={0 0 0},
}

\usepackage{colortbl}
\usepackage{multirow}

\newcommand{\figref}[1]{Fig.~\ref{#1}}

\newcommand{\tabref}[1]{Table~\ref{#1}}

\makeatletter
\DeclareRobustCommand\onedot{\futurelet\@let@token\@onedot}
\def\@onedot{\ifx\@let@token.\else.\fi}
\def\eg{e.g\onedot} 
\def\ie{i.e\onedot}

\makeatother

\definecolor{darkgreen}{rgb}{0,0.7,0}
\definecolor{darkblue}{RGB}{31,119,180}
\definecolor{darkred}{RGB}{214,39,40}
\definecolor{mediumgray}{rgb}{0.5,0.5,0.5}
\definecolor{mediumteal}{rgb}{0,0.5,0.5}
\definecolor{naviblue}{RGB}{0,0,128}

\definecolor{ellisred}{rgb}{0.87,0.44,0.38} %
\definecolor{ellisgreen}{rgb}{0.69,0.90,0.52} %
\definecolor{elliscyan}{rgb}{0.29,0.77,0.74} %
\definecolor{ellisorange}{rgb}{0.89,0.55,0.28} %
\definecolor{ellisblue}{rgb}{0.41,0.61,0.86} %

\definecolor{Tab0}{HTML}{1F77B4}
\definecolor{Tab1}{HTML}{ff7f0e}
\definecolor{Tab2}{HTML}{2ca02c}
\definecolor{Tab3}{HTML}{d62728}
\definecolor{Tab4}{HTML}{9467bd}
\definecolor{Tab5}{HTML}{8c564b}
\definecolor{Tab6}{HTML}{e377c2}
\definecolor{Tab7}{HTML}{7f7f7f}
\definecolor{Tab8}{HTML}{bcbd22}
\definecolor{Tab9}{HTML}{17becf}

\definecolor{Tabx0}{HTML}{4e79a7}
\definecolor{Tabx1}{HTML}{f28e2b}
\definecolor{Tabx2}{HTML}{e15759}
\definecolor{Tabx3}{HTML}{76b7b2}
\definecolor{Tabx4}{HTML}{59a14f}
\definecolor{Tabx5}{HTML}{edc948}
\definecolor{Tabx6}{HTML}{b07aa1}
\definecolor{Tabx7}{HTML}{ff9da7}
\definecolor{Tabx8}{HTML}{9c755f}
\definecolor{Tabx9}{HTML}{bab0ac}

\definecolor{textellisred}{HTML}{d85544}
\definecolor{textellisgreen}{HTML}{5fa02b}
\definecolor{textviolet}{HTML}{804c70}

\definecolor{customgray}{RGB}{136, 138, 133}

\newcommand{\rot}[1]{\multicolumn{1}{c}{\rotatebox{70}{#1}}}
\newcommand{\boldparagraph}[1]{\vspace{0.0cm}\noindent{\bf #1.} }

\definecolor{darkgreen}{rgb}{0,0.7,0}

\let\titleold\title
\renewcommand{\title}[1]{\titleold{#1}\newcommand{\thetitle}{#1}}

\newcommand{\trianglesymbol}[3][blue]{
  \begin{tikzpicture}[baseline=-0.6ex]
    \begin{scope}[rotate=#3, scale=#2]
      \filldraw[fill=#1, fill opacity=1.0, draw=black, line width=0.5pt] 
        (-0.5,-0.5) -- (0.75,0) -- (-0.5,0.5) -- (-0.25,0) -- cycle;
    \end{scope}
  \end{tikzpicture}
}

\title{Pseudo-Simulation for Autonomous Driving}

\author{
\textbf{Wei Cao}$^{*3,5}$ \quad
\textbf{Marcel Hallgarten}$^{*1,3,6}$ \quad
\textbf{Tianyu Li}$^{*4}$ \\
\textbf{Daniel Dauner}$^{1}$ \quad
\textbf{Xunjiang Gu}$^{6}$ \quad
\textbf{Caojun Wang}$^{4}$ \quad
\textbf{Yakov Miron}$^{3}$ \\
\textbf{Marco Aiello}$^{5}$ \quad
\textbf{Hongyang Li}$^{4}$ \quad
\textbf{Igor Gilitschenski}$^{6,7}$ \quad
\textbf{Boris Ivanovic}$^{2}$ \\
\textbf{Marco Pavone}$^{2,8}$ \quad
\textbf{Andreas Geiger}$^{1}$ \quad
\textbf{Kashyap Chitta}$^{1,2}$
\vspace{0.1cm}
\\
$^{1}$University of T{\"u}bingen, T{\"u}bingen AI Center \quad
$^{2}$NVIDIA Research \quad
$^{3}$Robert Bosch GmbH \\
$^{4}$OpenDriveLab at Shanghai Innovation Institute \quad
$^{5}$University of Stuttgart \\
$^{6}$University of Toronto \quad
$^{7}$Vector Institute \quad
$^{8}$Stanford University
}

\begin{document}
\maketitle

\enlargethispage{2\baselineskip}
{\let\thefootnote \relax \footnote{Primary contact: \texttt{kchitta@nvidia.com}. $^*$Equal contribution.}}

\setcounter{footnote}{0}

\begin{abstract}
Existing evaluation paradigms for Autonomous Vehicles (AVs) face critical limitations.
Real-world evaluation is often challenging due to safety concerns and a lack of reproducibility, whereas closed-loop simulation can face insufficient realism or high computational costs.
Open-loop evaluation, while being efficient and data-driven, relies on metrics that generally overlook compounding errors.
In this paper, we propose \textit{pseudo-simulation}, a novel paradigm that addresses these limitations.
Pseudo-simulation operates on real datasets, similar to open-loop evaluation, but augments them with synthetic observations generated prior to evaluation using 3D Gaussian Splatting.
Our key idea is to approximate potential future states the AV might encounter by generating a diverse set of observations that vary in position, heading, and speed.
Our method then assigns a higher importance to synthetic observations that best match the AV's likely behavior using a novel proximity-based weighting scheme.
This enables evaluating error recovery and the mitigation of causal confusion, as in closed-loop benchmarks, without requiring sequential interactive simulation.
We show that pseudo-simulation is better correlated with closed-loop simulations ($R^2=0.8$) than the best existing open-loop approach ($R^2=0.7$).
We also establish a public leaderboard for the community to benchmark new methodologies with pseudo-simulation.
Our code is available at  \href{https://github.com/autonomousvision/navsim}{{\color{ellisred}{\texttt{https://github.com/autonomousvision/navsim}}}}.
\end{abstract}
\section{Introduction}
\label{sec:introduction}

Reliable evaluation is essential for developing decision-making systems.
In the context of autonomous vehicles (AVs), this means assessing the system’s ability to navigate complex traffic scenarios efficiently, comfortably, and safely.
Existing evaluation strategies typically fall into two categories: closed-loop and open-loop evaluation~\cite{Chen2024PAMI}.

Closed-loop evaluation assesses model performance by placing it in an interactive environment.
The AV must safely navigate traffic while making progress toward a designated goal.
Although real-world closed-loop deployment offers reliable feedback, it is costly, risky, and not reproducible, making it insufficient on its own for benchmarking at the scale needed to demonstrate robustness~\cite{Buehler2009}.
As a more reproducible alternative, closed-loop evaluation is often conducted in simulation~\cite{Dosovitskiy2017CORL, Karnchanachari2024ICRA}.
Simulators enable rapid iteration and controlled scenario generation, and provide structured metrics for downstream performance analysis, such as collision or route completion rates.

However, accurate simulation remains a significant challenge, particularly for vision-based end-to-end AV systems.
Real-world driving is visually complex and behaviorally diverse, making it difficult to replicate in simulation.
Most existing platforms are manually constructed by 3D artists and engineers~\cite{Dosovitskiy2017CORL, Li2022PAMI}.
This limits their realism and the diversity across scenes.
Moreover, simulation-based evaluation is a computationally intensive and inherently sequential process. 
It often relies on large amounts of correlated evaluation frame sequences due to the high frequency of simulation required to ensure fidelity (usually 10Hz or higher).

Open-loop evaluation, on the other hand, measures planning performance by comparing predicted trajectories to expert demonstrations in pre-recorded datasets.
Each observation for evaluation includes sensor inputs, a goal location, and the future trajectory executed by a human expert driver.
The AV predicts a fixed-horizon trajectory conditioned on the inputs, which is then scored against the expert using either displacement errors or metrics derived from ground-truth (GT) environment annotations, such as lane compliance or estimated collisions~\cite{Weng2024CVPR, Li2024CVPR, Dauner2024NEURIPS}.
This approach operates entirely on real sensor data and avoids the complexities of interactive simulation, making it scalable and straightforward to apply over large datasets.
However, it evaluates behavior only under expert-aligned conditions and does not account for distribution shifts.
In deployment, the AV deviates from the demonstrated path, and open-loop protocols do not test its ability to recover from such drift.

\begin{figure}[t!]
    \centering
    \includegraphics[width=\textwidth]{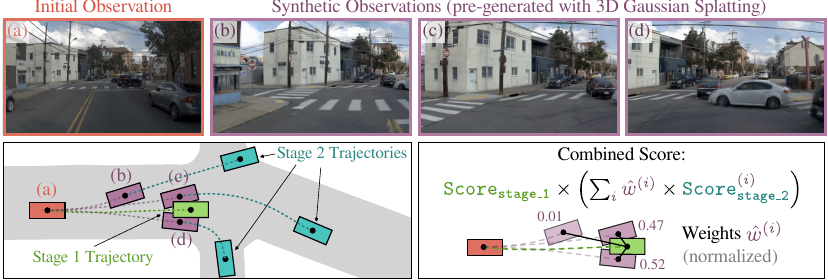}
    \caption{\textbf{Pseudo-simulation.} \textbf{(Top)} From an \textcolor{textellisred}{initial real-world observation (a)}, we generate \textcolor{textviolet}{synthetic observations (b, c, d)} via a variant of 3D Gaussian Splatting specialized for driving scenes~\cite{Li2025ARXIV}. Crucially, these synthetic observations are \textbf{pre-generated prior to evaluation}, unlike traditional interactive simulation where observations are generated online. \textbf{(Bottom)} Pseudo-simulation involves two stages. In Stage 1, we evaluate the AV's trajectory output for (a). Stage 2 involves evaluation on trajectories output for (b, c, d). Stage 2 scores are weighted ($\hat{w}^{(i)}$) based on the proximity of the \textcolor{textviolet}{Stage 2 synthetic observation start point} to the \textcolor{textellisgreen}{Stage 1 planned endpoint}. The aggregated score assesses robustness to small variations near the intended path, prioritizing the most likely futures.}
    \label{fig:teaser}
\end{figure}

To address the limitations of existing evaluation protocols, we introduce \textbf{pseudo-simulation}.
This new paradigm aims to combine the scalability of open-loop evaluation with a comprehensive assessment traditionally restricted to interactive closed-loop testing.
As shown in~\figref{fig:teaser}, our approach evaluates the AV's performance in two stages.
Stage 1 uses the originally recorded real-world observations.
Stage 2 uses synthetic observations generated based on these original frames.
{Crucially, these synthetic observations are generated before the evaluation process begins, enabling evaluation in a non-interactive manner.}
To generate Stage 2 observations, we adapt (to our data) a state-of-the-art driving scene reconstruction and rendering algorithm~\cite{Li2025ARXIV} based on 3D Gaussian Splatting~\cite{Kerbl2023TOG}.

We evaluate the output trajectories predicted by the AV, considering its performance on both the initial real-world observations (from Stage 1) and the generated synthetic observations (used in Stage 2).
Our key idea lies in how we assess performance in Stage 2: we weight the importance of each synthetic observation based on its proximity to the endpoint of the trajectory that the AV initially predicted in Stage 1 (\figref{fig:teaser} bottom-right).
This weighting strategy allows the evaluation to better reflect the AV's robustness and ability to recover from potential errors when facing conditions similar to what it may encounter in a closed-loop simulation.
Our approach also inherently assigns lower weights to synthetic observations that significantly deviate from the initially predicted endpoint, thereby preventing undue penalties for failures in improbable or irrelevant future states.

\begin{wraptable}{r}{7.8cm}
\centering
\scriptsize
\renewcommand{\arraystretch}{1.2}
\setlength{\tabcolsep}{3.5pt}
\rowcolors{2}{white}{ellisblue!20}
\vspace{0.0cm}
\begin{tabular}{>{\columncolor{ellisblue!20}}l 
                >{\columncolor{ellisblue!20}}c 
                >{\columncolor{ellisblue!20}}c 
                >{\columncolor{ellisblue!20}}c}
\rowcolor{naviblue}
\textcolor{white}{\textbf{Property}} &
\textcolor{white}{\textbf{Open-Loop}} & 
\textcolor{white}{\textbf{Closed-Loop}} & 
\textcolor{white}{\textbf{Pseudo-Sim}} \\
\textbf{Real data} & \cellcolor{ellisgreen!20}Yes & \cellcolor{ellisred!20}No & \cellcolor{ellisgreen!20}Yes \\
\textbf{Per-scene pre-processing} & \cellcolor{ellisgreen!20}No & \cellcolor{ellisred!20}Yes & \cellcolor{ellisred!20}Yes \\
\textbf{Synthetic data rendering} & \cellcolor{ellisgreen!20}No & \cellcolor{ellisred!20}Online & \cellcolor{ellisgreen!20}Pre-rendered \\
\textbf{Evaluation} & \cellcolor{ellisgreen!20}Parallel & \cellcolor{ellisred!20}Sequential & \cellcolor{ellisgreen!20}Parallel \\
\textbf{Compounding errors} & \cellcolor{ellisred!20}No & \cellcolor{ellisgreen!20}Yes & \cellcolor{ellisgreen!20}Yes \\
\textbf{Causal confusion} & \cellcolor{ellisred!20}No & \cellcolor{ellisgreen!20}Yes & \cellcolor{ellisgreen!20}Yes \\
\end{tabular}
\vspace{-0.6cm}
\end{wraptable}

In summary, pseudo-simulation combines \textbf{{real}} and \textbf{pre-rendered synthetic} data, enabling \textbf{scalable, parallel} evaluation.

We show that pseudo-simulation achieves strong correlation with closed-loop results for a set of 83 diverse planners nuPlan~\cite{Karnchanachari2024ICRA}, while being substantially more efficient (6$\times$ less environment interactions).
To enable standardized benchmarking, we release NAVSIM v2, a framework for benchmarking autonomous driving built upon our proposed evaluation methodology.
We find that it reveals previously unknown failure modes in popular AV algorithms~\cite{Chitta2023PAMI,Dauner2023CORL}, thus establishing it as a challenging new testbed for future research.
We hope that pseudo-simulation can accelerate AV development through more efficient experimentation cycles and ensuring that future models prioritize closed-loop robustness.

\section{Related Work}
\label{sec:related}

\boldparagraph{Counterfactual Data Augmentation}
Counterfactual augmentation has been used to expose models to out-of-distribution data by generating structured perturbations~\cite{PITIS2020NIPS, PITIS2022NIPS, CHEN2021ACL}.
Related work for AVs focuses on augmenting training data with viewpoint shifts~\cite{Bojarski2016ARXIV, CHEN2020CORL, Codevilla2019ICCV, Prakash2020CVPR, Jaeger2023ICCV, MA2024ARXIV, ZIMMERLIN2024ARXIV, RENZ2025ARXIV}. We make the first attempt to adopt such augmentations primarily for {evaluation}.

\boldparagraph{Closed-Loop Benchmarking}
Graphics-based simulators support closed-loop evaluation, but are computationally expensive and introduce domain gaps in sensor fidelity~\cite{Dosovitskiy2017CORL,Li2022PAMI}.
To improve scalability, data-driven planning simulators leverage recorded traffic data~\cite{Karnchanachari2024ICRA, Gulino2023NIPS, Vinitsky2022NIPS, Kazemkhani2024ARXIV}.
However, these systems operate at the trajectory level and do not support sensor-based agents.
Several works attempt to bridge this gap through data-driven sensor simulation, generating synthetic views from real-world logs.
Early systems simulate ego-vehicle deviations via image-based rendering~\cite{Amini2020RAL, Amini2022ICRA, Wang2022ICRA}.
More recent methods explore neural rendering~\cite{Yang2023CVPR, Tonderski2024CVPR, Ljungbergh2024ECCV, Yang2024ARXIV, Zhou2024ARXIV, You2024ARXIV}.
However, they face challenges related to photorealism and runtime efficiency.
As such, there is no widely established neural rendering based AV benchmark yet, which we aim to address in this work.

\boldparagraph{Open-Loop Benchmarking}
Open-loop evaluation typically measures planning quality via displacement errors between predicted and expert trajectories~\cite{Sima2024ECCV}.
These metrics are simple to compute, but often correlate poorly with real-world performance and tend to favor trivial or history-based baselines~\cite{Li2024CVPR, Dauner2023CORL, Codevilla2018ECCV, Zhai2023ARXIV}.
Furthermore, benchmarks adopting the nuScenes dataset~\cite{Caesar2021CVPR} exhibit inconsistencies in implementations for metrics such as ADE (Average Displacement Error) and collisions~\cite{Weng2024CVPR, Li2024CVPR}, and overrepresent low-complexity scenes such as straight driving~\cite{Li2024CVPR}.
The closest work to ours, NAVSIM v1~\cite{Dauner2024NEURIPS}, offers a more structured framework for open-loop benchmarking.
The agent-under-test predicts a fixed-horizon trajectory from real sensor inputs, while other actors replay their recorded motion.
This setup supports scalable evaluation and enables simulation-based metrics such as progress and collision rates.
However, unlike pseudo-simulation, NAVSIM v1 remains limited to open-loop evaluation from expert-aligned initial observations and does not account for compounding errors or causal confusion~\cite{Wen2020NEURIPS}.
\section{Pseudo-Simulation}
\label{sec:method}

We consider a planning task where evaluation proceeds in two stages (\figref{fig:teaser}). In both stages, an AV (also called planner/ego agent) generates a 4-second trajectory based on sensor inputs and a driving command~\cite{Dauner2024NEURIPS}.
The inputs include multi-view camera images and ego status features such as the velocity and motion history.
The driving command specifies the intended maneuver in case of ambiguity, \eg, at intersections, and is provided as a discrete label: \textit{left}, \textit{straight}, or \textit{right}.
The ego agent outputs a trajectory (\ie, a sequence of desired future waypoints) in its local coordinate frame.

\subsection{Stage 1: Initial Observations}
\label{sec:method_1}

In Stage 1, we infer the ego agent's motion based on an initial observation from the test dataset.
We then simulate a simplified Bird's Eye View (BEV) representation of the scene forward for a fixed time horizon, obtaining a score as well as an endpoint to be used later in Stage 2.

\boldparagraph{BEV Simulation}
The 4-second trajectory predicted by the agent is executed using a kinematic bicycle model~\cite{Rajamani2011Springer} and an LQR controller~\cite{Lehtomaki1981TAC} at 10Hz.
The trajectory is committed for the entire simulation horizon, and no closed-loop feedback is provided to the agent during this time.
Unlike related prior work~\cite{Dauner2024NEURIPS}, which uses non-reactive traffic to simplify implementation (\ie, neighboring vehicles follow their recorded trajectories without reacting to the ego agent), we improve the simulation realism with reactive traffic.
Background vehicles (represented as oriented bounding boxes) respond to the ego agent using a rule-based planner called the Intelligent Driver Model (IDM)~\cite{Treiber2000}.
Pedestrians, static obstacles, and other non-vehicle actors follow their recorded trajectories without reacting to the ego agent.

\boldparagraph{Extended PDM Score}
Our metric, the Extended Predictive Driver Model Score (EPDMS)~\cite{Li2025ARXIVb}, builds on the PDMS introduced in prior work~\cite{Dauner2024NEURIPS}.
Besides minor modifications (detailed in the supplementary material), the design of the metric is largely consistent with~\cite{Li2025ARXIVb}.
It combines multiplicative penalties for rule violations with a weighted average of several subscores:
\vspace{-0.0cm}
\begin{equation}
\text{EPDMS} = 
\underbrace{
\prod_{m \in \mathcal{M}_\text{pen}} \text{filter}_m(\text{agent}, \text{human})
}_{\text{penalty terms}}
\cdot
\underbrace{
\frac{ \sum_{m \in \mathcal{M}_\text{avg}} w_m \cdot \text{filter}_m(\text{agent}, \text{human}) }
     { \sum_{m \in \mathcal{M}_\text{avg}} w_m }
}_{\text{weighted average terms}}
\label{eq:epdms}
\end{equation}

\begin{wraptable}{r}{7.2cm}
\small
    \vspace{-0.6cm}
    \centering
    \setlength{\tabcolsep}{3pt}
    \begin{tabular}{l|c|c}
        \toprule
        \textbf{Subscore} & \textbf{$w_m$} & \textbf{Range} \\
        \midrule
        No at-fault Coll. (NC) & - & $\{0,\frac{1}{2},1\}$ \\ 
        Drivable Area Compl. (DAC) & - & $\{0,1\}$ \\ 
        Driving Direction Compl. (DDC) & - & $\{0,\frac{1}{2},1\}$ \\ 
        Traffic Light Compl. (TLC) &  - & $\{0,1\}$ \\ 
        \cmidrule(l{5pt}r{5pt}){1-3}
        Ego Progress (EP) & 5 & $[0,1]$ \\ 
        Time to Collision (TTC) & 5 & $\{0,1\}$ \\ 
        Lane Keeping (LK) & 2 & $\{0,1\}$ \\ 
        History Comfort (HC) & 2 & $\{0,1\}$ \\ 
        Extended Comfort (EC) & 2 & $\{0,1\}$ \\ 
        \bottomrule
    \end{tabular}
    \caption{\textbf{EPDMS.} Subscores, weights, and ranges.}
    \label{tab:epdms_weights}
    \vspace{-0.0cm}
\end{wraptable}

Here, $\mathcal{M}_\text{pen} = \{\text{NC}, \text{DAC}, \text{DDC}, \text{TLC}\}$ and $\mathcal{M}_\text{avg} = \{\text{TTC}, \text{EP}, \text{HC}, \text{LK}, \text{EC}\}$ (\tabref{tab:epdms_weights}).
Unlike prior work~\cite{Li2025ARXIVb}, to prevent penalizing contextually justified maneuvers, we introduce a novel filtering mechanism ($\text{filter}_m$) for the EPDMS.
If a rule violation is also committed by the human expert driver in the same scene, the penalty is ignored.
This avoids penalizing infractions due to label noise or valid behaviors, such as briefly entering the opposite lane to bypass a static obstacle.

\subsection{Stage 2: Synthetic Observations}
\label{sec:method_2}

\begin{figure}[t!]
    \centering
    \includegraphics[width=0.4971\textwidth]{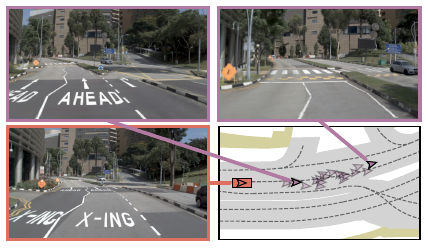}
    \includegraphics[width=0.4971\textwidth]{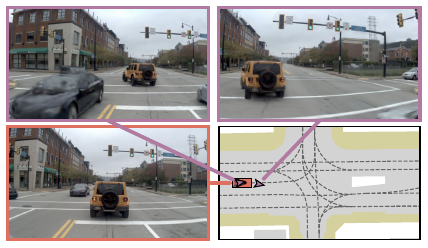}
    \includegraphics[width=0.4971\textwidth]{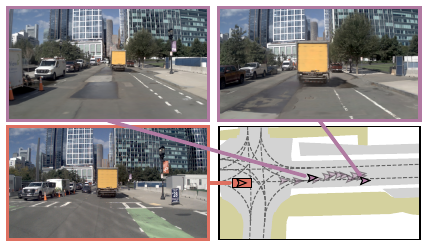}
    \includegraphics[width=0.4971\textwidth]{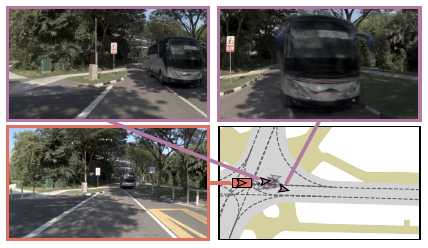}
    \caption{\textbf{Example scenes.} We show the poses and front-view camera images for the \textcolor{textellisred}{initial real-world observation} (\trianglesymbol[ellisred]{0.2}{0}) and \textcolor{textviolet}{pre-generated synthetic observations} (\trianglesymbol[Tabx6]{0.2}{0}) in four scenes.}
    \label{fig:navhard}
    \vspace{-0.5cm}
\end{figure}

In Stage 2, the agent's behavior is inferred on pre-generated synthetic observations.
The scoring pipeline from Stage 1 is repeated for each of these synthetic observations.
Stage 2 scores correspond to a range of plausible futures.
We propose to weight their contributions towards a final combined score based on the proximity of Stage 2 start points to the Stage 1 endpoint.
This prioritizes futures that are more likely.
We show some examples of such generated scenes in \figref{fig:navhard}.
In the following, we provide details regarding the scenario pre-generation, scoring, and score aggregation processes.
Note that we choose to create synthetic observations after unrolling for 4 seconds, instead of directly at the Stage 1 start point, since (1) this allows background traffic to react to the updated ego state, and (2) it provides a physically plausible history trajectory, which is a required planner input.

\boldparagraph{Start Point Sampling}
As a data pre-processing step prior to the evaluation of any specific planner, we generate Stage 2 synthetic observations that approximate the range of possible rollout endpoints for Stage 1 observations in the dataset.
Each Stage 2 observation must have a valid start point and heading, with an associated motion history, and multi-view camera image inputs for a planner.

We sample start points around the {expert driver}'s observed endpoint after 4 seconds in the scene.
Importantly, this sampling does not depend on the Stage 1 endpoint produced by a planner, but only the expert driving trajectories from the original dataset, available prior to evaluation.
We define a sampling region around this expert endpoint: laterally, viewpoints are sampled every 0.5 meters up to 2.0 meters on each side; longitudinally, viewpoints are sampled every 5.0 meters.
The longitudinal sampling spans the physically plausible range from the minimum stopping distance to the maximum reachable distance (assuming accelerations of $\pm$4.0 m/s$^2$ for 4 seconds).
This naturally produces more potential states for high-speed scenarios (up to 20 in practice) compared to low-speed ones.

\boldparagraph{Heading and History Generation}
For each sampled start point, we generate a plausible heading and motion history by matching it to the nearest trajectory in a human driving dataset.
This matching process includes filtering: we discard candidate trajectories if they differ in velocity by more than 1.0 m/s, acceleration by more than 1.0 m/s$^2$, or heading by more than 20 degrees relative to the expert.
We then apply rejection sampling to remove any remaining start points that violate the multiplicative EPDMS constraints (NC, DAC, DDC and TLC).
Finally, we discard scenes from the neural reconstruction pipeline if fewer than five valid synthetic observations remain after filtering.

\boldparagraph{Neural Reconstruction and Rendering} 
We employ a state-of-the-art dynamic scene reconstruction approach to achieve high-fidelity neural rendering. Specifically, we use a modified version of Multi-Traversal Gaussian Splatting (MTGS)~\cite{Li2025ARXIV}.
As in MTGS, we model scene dynamics with a scene graph~\cite{Ost2021CVPR}.
However, unlike MTGS, which uses multiple nearby driving traversals for jointly optimizing a 3D scene representation, we use only a single traversal.
This significantly expands the pool usable data, as only a subset of our dataset includes multiple co-located traversals.
To reduce localization noise, we calculate accurate initial camera pose estimates via LiDAR registration~\cite{Vizzo2023RAL} and bundle adjustment~\cite{Schoenberger2016CVPR}, followed by camera pose optimization during the MTGS training process~\cite{Yan2024CVPR}.
Before reconstruction, we filter out scenes affected by significant sensor failures (water droplets or flares).
After reconstruction, we apply a semi-automatic filtering step to discard reconstructed scenes of low visual quality (details in supplementary material).

\boldparagraph{Score Aggregation}  
We score each synthetic observation using the EPDMS from \Cref{eq:epdms} to obtain Stage 2 scores \(\{s_2^i\}\).
To compute the final score $s_{\text{combined}}$, we define two aggregation functions:
$
s_{\text{combined}} = \mathcal{A}_{\text{1}}(s_1, s_2), \text{where} \
s_2 = \mathcal{A}_{\text{2}}(\{s_2^i\}, \{x^i\}, \hat{x})
$.
\(\mathcal{A}_{\text{1}}\) fuses the Stage 1 score \(s_1\) with an aggregated Stage 2 result \(s_2\).
\(\mathcal{A}_{\text{2}}\), in turn, aggregates \(\{s_2^i\}\) based on their initial positions \(\{x^i\}\), which denote the start points of the \(i\)-th Stage 2 scenario.
\(\hat{x}\) is the ego agent's endpoint reached at the end of the Stage 1 simulation.  
In our experiments, we conduct an empirical study on different aggregation functions.
Based on our findings, we instantiate \(\mathcal{A}_{\text{1}}\) as a simple product, and \(\mathcal{A}_{\text{2}}\) as a Gaussian-weighted average with kernel variance $\sigma^2$:
\begin{equation}
s_{\text{combined}} = s_1s_2,
\quad
s_2 = \sum_i \hat{w}^i s_2^i, 
\quad
\hat{w}^i = \frac{w^i}{\sum_i w^i}, 
\quad
w^i = \exp\left(-\frac{\|x^i - \hat{x}\|^2}{2\sigma^2}\right)
\label{eq:agg_2}
\end{equation}
\section{Results}
\label{sec:results}

In this section, we study three key questions to examine our proposed idea of pseudo-simulation.

\subsection{How well-aligned is pseudo-simulation with closed-loop evaluation?}
\label{sec:results_correlation}

\begin{figure}[t!]
    \centering
    \includegraphics[width=\textwidth]{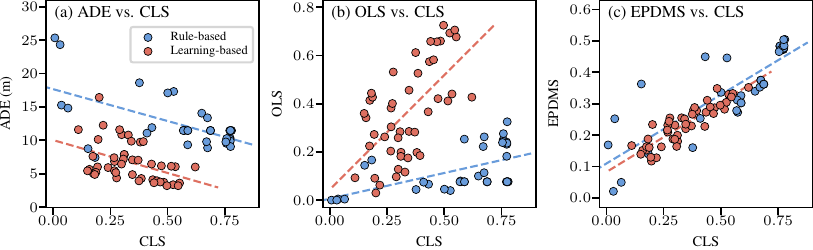}
    \caption{\textbf{Pseudo-simulation correlates strongly with closed-loop evaluation.} Compared to the average displacement error (ADE) and nuPlan open-loop score (OLS) metrics, EPDMS achieves a strong correlation to the closed-loop score (CLS) for a set of 37 rule-based and 46 learned planners.}
    \label{fig:correlation_scatter}
\end{figure}

\boldparagraph{Benchmark}
To evaluate how well pseudo-simulation aligns with closed-loop simulation, we conduct a correlation analysis with the nuPlan simulator~\cite{Karnchanachari2024ICRA}.
nuPlan supports fully reactive rollouts for privileged planners with access to ground-truth perception and HD maps.
We include a total of 83 planners, comprising both rule-based and learned models, to represent a wide range of behaviors and performance levels.
For rule-based methods, we use 10 constant kinematics baselines, 15 IDM planners~\cite{Treiber2000}, and 15 PDM-Closed variants~\cite{Dauner2023CORL}.
For learned approaches, we evaluate 22 PlanCNN~\cite{Renz2022CORL} models with varying input modalities and 24 Urban Driver~\cite{Scheel2021CORL} models differing in architecture and training configurations.
For these experiments, we use a reduced version of EPDMS, excluding the TLC, LK, and EC metrics, because nuPlan does not support these for closed-loop evaluation.

We measure the alignment between EPDMS and nuPlan's closed-loop score (CLS) using Pearson's linear ($r$) and Spearman's rank ($\rho$) correlation coefficients, as well as the coefficient of determination ($R^2$).
Since $R^2$ is calculated by fitting a linear model between EPDMS and CLS, it is equivalent to the square of Pearson's correlation coefficient here ($R^2 = r^2$).
This assumes that an ideal pseudo-simulation metric should show a linear relationship with closed-loop scores, requiring no adjustments for scale or bias.
We evaluate each planner on a filtered subset of nuPlan, described in detail in the supplementary material, to collect both closed-loop and pseudo-simulation scores.
This subset includes 244 initial observations (Stage 1) and 4164 synthetic observations (Stage 2).

\boldparagraph{Results}
First, we create a scatter plot comparing the 8-second closed-loop scores from the nuPlan simulator against the results of our 2$\times$4 second pseudo-simulation, which aims to approximate these closed-loop scores.
As shown in~\figref{fig:correlation_scatter}, pseudo-simulation exhibits strong correlation with closed-loop results across a broad range of planners, particularly among learned planners.
In contrast, prior open-loop evaluation metrics such as average displacement error and nuPlan's open-loop score~\cite{Karnchanachari2024ICRA} show nearly no correlation to closed-loop evaluation, consistent with the findings of~\cite{Dauner2024NEURIPS}.

\begin{figure}[t!]
    \centering
    \includegraphics[width=\textwidth]{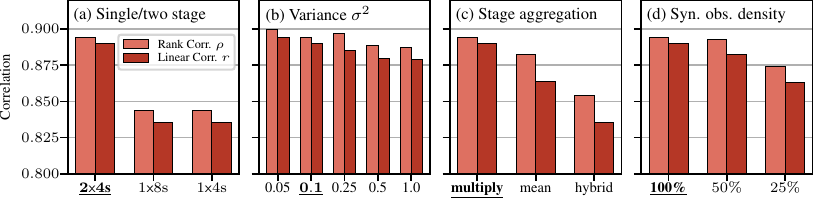}
    \caption{\textbf{EPDMS design choices.} We compare the correlation between the pseudo-simulation metric and the closed-loop score (CLS) under several settings. (a) single ($1\mathsf{x}$) vs. two stage ($2\mathsf{x}$) evaluation, (b) Gaussian weight variances, (c) Stage 1 and 2 aggregation methods, and (d) synthetic observation densities. Defaults in \underline{\textbf{bold-underline}}.}
    \label{fig:correlation}
\end{figure}

In~\figref{fig:correlation} (a), we compare single-stage open-loop simulation (at 4 and 8 seconds) to our two-stage pseudo-simulation variant. The two-stage setup achieves significantly higher alignment, reaching a Pearson correlation of $r = 0.89$ (corresponding to $R^2 = 0.8$), compared to $r=0.83$ ($R^2 = 0.7$) for the single-stage baselines.
Furthermore, compared to standard reactive closed-loop evaluation, our pseudo-simulation method exposes a wider range of potential failures.
This typically results in lower average EPDMS values compared to CLS values.
By injecting synthetic deviations, pseudo-simulation effectively reveals edge cases that might not be encountered during standard testing.

Within Stage 2, we assess the impact of the weighting used to combine scores across synthetic viewpoints.
~\figref{fig:correlation} (b) shows the correlation with closed-loop scores for different kernel variances.
We observe that smaller variances lead to improved results.
$\sigma^2 = 0.05$ and our default configuration of $\sigma^2 = 0.1$ give the highest correlations.
In additional experiments (included in the supplementary material), we find that other approaches, such as simple averaging, $k$-nearest neighbors ($k$-NN), and hybrid $k$-NN/Gaussian weighting are less effective than our default configuration.

To combine metrics across stages, we compare multiplicative aggregation, aggregation by the arithmetic mean, and a hybrid strategy where penalty metrics (e.g., collision and drivable area compliance) are multiplied while the remaining terms are averaged.
~\figref{fig:correlation} (c) summarizes the correlation of each strategy with closed-loop scores.
Multiplicative aggregation shows both higher linear and rank correlation than the other approaches.
This outcome is likely because most subscores are binary (i.e., 0 or 1). Consequently, multiplication appears to be a more suitable method for estimating the overall score for an 8-second interval based on two 4-second segments.

Finally, we examine the effect of limiting the number of synthetic views in Stage 2.
~\figref{fig:correlation} (d) reports correlation values when using 100\%, 50\%, and 25\% of our available synthetic viewpoints.
At 100\% density, each scenario contains 12 synthetic observations on average in Stage 2 for each real observation in Stage 1, resulting in 13 planner inferences per scenario.
In comparison, closed-loop simulation in nuPlan requires 80 planner inferences per scenario, corresponding to an 8-second rollout at 10Hz.
This is 6$\times$ higher than pseudo-simulation.
While subsampling reduces the number of synthetic views, the correlation to closed-loop scores remains strong.
Even when using only 25\% density, \ie, approximately three Stage 2 observations per scene, the correlation remains above 0.85.
This indicates that pseudo-simulation maintains reliability even with reduced observation coverage.

\subsection{What new challenges and insights does our leaderboard provide?}
\label{sec:results_leaderboard}

\begin{table}[t!]
\small
\setlength{\tabcolsep}{5pt} %
    \begin{center}
    \resizebox{0.98\textwidth}{!}{%
            \begin{tabularx}{\textwidth}{c c | YYYY | YYYYYY | c} 
                \toprule
                & \textbf{Stage} & \rot{CV~\cite{Dauner2024NEURIPS}} & \rot{Ego MLP~\cite{Dauner2024NEURIPS}} & \rot{LTF~\citep{Chitta2023PAMI}} & \rot{NavFormer} & \rot{LTFv6~\cite{Nguyen2026CVPR}} & \rot{RAP~\cite{Feng2025ARXIV}} & \rot{ZTRS~\cite{Li2025ARXIVc}} & \rot{GuideFlow~\cite{Liu2025ARXIV}} & \rot{SimScale~\cite{Tian2026CVPR}} & \rot{DrivoR~\cite{Kirby2026CVPR}} & \rot{\textit{PDM-C~\cite{Dauner2023CORL}}} \\
                \midrule
                \multirow{2}{*}{\textbf{NC} $\uparrow$} & S1 & 88.8 & 93.2 & 96.2 & 96.2 & 96.5 & 97.1 & 98.8 & \textbf{99.5} & \textbf{99.5} & 99.1 & \textit{94.4} \\
                 & S2 & 83.2 & 77.2 & 77.7 & 85.7 & 79.8 & 83.2 & 91.1 & 91.4 & \textbf{94.5} & 92.3 & \textit{90.5} \\
                 \cmidrule{2-13}
                \multirow{2}{*}{\textbf{DAC} $\uparrow$} & S1 & 42.8 & 55.7 & 79.5 & 92.4 & 86.6 & 94.4 & 97.5 & 98.0 & \textbf{99.1} & 98.2 & \textit{98.8} \\
                 & S2 & 59.1 & 51.9 & 70.2 & 81.0 & 75.5 & 83.8 & 90.4 & 89.5 & \textbf{94.2} & 91.6 & \textit{90.6} \\
                 \cmidrule{2-13}
                \multirow{2}{*}{\textbf{DDC} $\uparrow$} & S1 & 70.6 & 86.6 & 99.1 & 95.7 & 99.2 & 98.7 & \textbf{100.0} & 99.4 & 99.8 & 99.3 & \textit{100} \\ 
                 & S2 & 76.5 & 74.4 & 84.2 & 83.5 & 86.2 & 87.3 & 95.7 & 95.1 & 95.7 & \textbf{97.3} & \textit{95.4} \\
                 \cmidrule{2-13}
                \multirow{2}{*}{\textbf{TLC} $\uparrow$} & S1 & 99.3 & 99.3 & 99.5 & 99.6 & 99.5 & 99.7 & \textbf{100.0} & 99.3 & \textbf{100.0} & 99.7 & \textit{99.5} \\
                 & S2 & 98.0 & 98.2 & 98.0 & 97.6 & 97.8 & 98.0 & 99.0 & 98.8 & \textbf{99.2} & 99.0 & \textit{98.4} \\
                \midrule
                \multirow{2}{*}{\textbf{EP} $\uparrow$} & S1 & 77.5 & 81.2 & 84.1 & 83.8 & \textbf{84.4} & 83.8 & 66.6 & 79.6 & 69.6 & 75.4 & \textit{100} \\
                 & S2 & 71.3 & 77.1 & 85.1 & \textbf{90.1} & 89.5 & 86.8 & 63.5 & 77.5 & 75.8 & 75.7 & \textit{100} \\
                 \cmidrule{2-13}
                \multirow{2}{*}{\textbf{TTC} $\uparrow$} & S1 & 87.3 & 92.2 & 95.1 & 96.0 & 95.1 & 96.8 & 98.8 & 99.3 & \textbf{99.5} & 98.6 & \textit{93.5} \\
                 & S2 & 81.1 & 75.0 & 75.6 & 82.4 & 76.0 & 80.3 & 89.7 & 89.5 & \textbf{92.8} & 90.5 & \textit{86.6} \\
                 \cmidrule{2-13}
                \multirow{2}{*}{\textbf{LK} $\uparrow$} & S1 & 78.6 & 83.5 & 94.2 & 94.7 & 94.4 & 94.6 & \textbf{96.2} & 94.8 & 95.7 & 94.8 & \textit{99.3} \\
                 & S2 & 47.9 & 40.8 & 45.4 & 48.2 & 50.0 & 52.2 & \textbf{60.4} & 52.5 & 60.0 & 56.0 & \textit{74.2} \\
                 \cmidrule{2-13}
                \multirow{2}{*}{\textbf{HC} $\uparrow$} & S1 & 97.1 & 97.5 & 97.5 & 96.4 & \textbf{97.7} & 96.4 & 96.6 & 97.1 & 95.5 & 97.5 & \textit{87.7} \\
                 & S2 & 97.1 & 97.8 & 95.7 & 94.9 & 95.2 & 95.1 & 97.6 & 93.5 & 96.0 & \textbf{98.4} & \textit{91.9} \\
                 \cmidrule{2-13}
                \multirow{2}{*}{\textbf{EC} $\uparrow$} & S1 & 60.4 & 77.7 & \textbf{79.1} & 60.9 & {76.4} & 66.2 & 44.0 & 58.2 & 28.4 & 70.2 & \textit{36.0} \\
                 & S2 & 61.9 & \textbf{79.8} & 75.9 & 48.4 & 66.7 & 52.4 & 66.0 & 51.0 & 43.2 & 44.7 & \textit{29.7} \\
                \midrule
                    \multicolumn{2}{c|}{\textbf{EPDMS} $\uparrow$}   & 11.4 & 14.1 & 25.1 & 34.1 & 31.9 & 39.6 & 48.1 & 51.5 & 53.2 & \textbf{54.5} & \textit{56.6} \\ 
                \bottomrule
            \end{tabularx}%
            }
    \end{center}
    \caption{\textbf{\texttt{navhard} leaderboard.} Snapshot from 03/2026.}
    \label{tab:sota_transposed}
\end{table}

\boldparagraph{Benchmark} Our public NAVSIM v2 leaderboard\footnote{\href{https://huggingface.co/spaces/AGC2025/e2e-driving-navhard}{{\color{ellisred}{\texttt{https://huggingface.co/spaces/AGC2025/e2e-driving-navhard}}}}} features challenging driving scenarios (\eg~unprotected turns and dense traffic, see~\figref{fig:navhard}).
It uses a subset of nuPlan that we refer to as \texttt{navhard}, involving 450 Stage 1 and 5462 Stage 2 observations.
The goal of the leaderboard is to ensure standardized evaluation, as subtle differences (\eg, enabling/disabling the human-based filtering mechanism, or using a newer \texttt{numpy} version) can have an effect on the computed metrics.
Therefore, we discourage the use of self-reported and unofficial ``NAVSIM v2'' benchmark splits, such as reporting the EPDMS on the NAVSIM v1 \texttt{navtest} dataset without conducting two-stage pseudo-simulation. Instead, submission to our leaderboard can ensure both consistency and visibility of submissions.

For our analysis, we select five baseline planners with varying input modalities, as well as six community-contributed submissions: LTFv6~\cite{Nguyen2026CVPR}, RAP~\cite{Feng2025ARXIV}, ZTRS~\cite{Li2025ARXIVc}, GuideFlow~\cite{Liu2025ARXIV}, SimScale~\cite{Tian2026CVPR}, and DrivoR~\cite{Kirby2026CVPR}.
Our baselines include the Constant Velocity (CV) and Ego-history MLP from~\cite{Dauner2024NEURIPS}, Latent TransFuser (LTF)~\cite{Chitta2023PAMI}, the most established baseline for image-only planning for nuPlan, and PDM-Closed (PDM-C)~\cite{Dauner2023CORL}, the best privileged planner on nuPlan.
To tackle the complex scene types and visual perturbations of \texttt{navhard}, we propose a new simple and robust baseline architecture called {NavFormer}\footnote{\href{https://github.com/OpenDriveLab/NavFormer}{{\color{ellisred}{\texttt{https://github.com/OpenDriveLab/NavFormer}}}}}. 
This model employs a BEVFormer encoder~\cite{Li2022ECCV} to process surround-view camera inputs, in contrast to LTF which only encodes forward-facing camera data.
Following PARA-Drive~\cite{Weng2024CVPR}, object tracking and map segmentation decoders provide auxiliary perception supervision.
Finally, NavFormer uses a Hydra-MDP decoder head for planning~\cite{Li2024ARXIV}.

\boldparagraph{Results}
Table~\ref{tab:sota_transposed} presents the detailed subscores for each planner, broken down by Stage 1 (original observations) and Stage 2 (synthetic observations).
In terms of overall performance, PDM-Closed achieves the highest EPDMS of 56.6, closely followed by recent scalable architectures like DrivoR~\cite{Kirby2026CVPR}.
Comparing performance across stages, we observe a general drop in subscores from Stage 1 to Stage 2 for all methods, suggesting their sensitivity to the distribution shifts introduced in Stage 2.
Notably, while PDM-Closed excels in most metrics, it exhibits lower performance in comfort metrics such as HC and EC.
Our evaluation on \texttt{navhard} reveals this specific failure mode of PDM-Closed, highlighting a trade-off that was overlooked in prior benchmarks.

\subsection{Does the proposed neural rendering yield sufficient visual fidelity?}
\label{sec:results_rendering}

\boldparagraph{Benchmark}
To assess the fidelity of our synthetic observations, we evaluate the impact of our neural rendering on the downstream perception and planning performance of a pre-trained model using the \texttt{navhard} dataset.
Specifically, we employ the LTF model from~\tabref{tab:sota_transposed}.
Although primarily an end-to-end planner, LTF outputs intermediate Bird's Eye View (BEV) segmentations and, crucially, is trained only on real-world data.
This allows us to measure the domain gap introduced by our rendering: we evaluate LTF's performance on synthetic data and compare it to its performance on real data.
We use mean Intersection over Union (mIoU) over the drivable area, walkway, and vehicle classes output by LTF to evaluate BEV perception, and the EPDMS metric to evaluate planning.

\boldparagraph{Results}
We present our findings in~\tabref{tab:rendering_1}.
First, we evaluate perception performance using the LTF model. Comparing Stage 1 to Stage 2 views, we observe a drop in mIoU from 46.0 to 37.6.
Despite this degradation in segmentation quality, planning performance remains largely stable, with EPDMS decreasing only slightly from 62.3 to 61.0.
While mIoU captures semantic segmentation fidelity, it does not directly reflect planner-relevant errors~\cite{Behl2020IROS, Schreier2023ICCVW}.
For our data, the observed reduction in mIoU does not appear to impair semantic cues needed for planning.
This suggests that our synthetic observations preserve the most critical information.

Next, we evaluate performance under perturbed synthetic Stage 2 inputs.
Here, mIoU drops marginally from 37.6 to 36.9, while EPDMS declines substantially to 44.2.
This larger drop is consistent with trends observed previously (Section~\ref{sec:results_leaderboard}), where non-privileged planners showed greater sensitivity to deviations from expert trajectories.
The small change in mIoU between the synthetic settings of Stage 1 and Stage 2, compared to the greater drop in EPDMS, suggests that the observed planning degradation is primarily driven by the planner's sensitivity to the distribution shift, rather than by perception inaccuracies stemming from rendering artifacts.

\begin{table}[t!]
\small
    \begin{tabular}{cc}
        \begin{subtable}[b]{0.6\textwidth}
                \begin{tabular}{lc|c|c}
                    \toprule
                    \textbf{Data} & \textbf{Stage} & \textbf{Perception mIoU} $\uparrow$ & \textbf{Planning EPDMS} $\uparrow$ \\
                    \midrule
                    Real & S1 & 46.0 & 62.3 \\
                    Syn. & S1 & 37.6 & 61.0 \\
                    \midrule
                    Syn. & S2 & 36.9 & 44.2 \\
                    \bottomrule
                \end{tabular}
            \caption{\textbf{Synthetic data quality for downstream tasks.}}
            \label{tab:rendering_1}
        \end{subtable}
        &
        \begin{subtable}[b]{0.3\textwidth}
                \begin{tabular}{l|cc|cc}
                    \toprule
                    \textbf{Method} & \textbf{LPIPS} $\downarrow$ \\
                    \midrule
                    Street Gaussians~\cite{Yan2024ECCV} & 0.354 \\
                    Ours (w/o pose opt.) & 0.322 \\
                    \midrule
                    Ours & 0.253 \\
                    \bottomrule
                \end{tabular}
            \caption{\textbf{NVS ablation study.} }
            \label{tab:rendering_2}
        \end{subtable}
    \end{tabular}
    \caption{\textbf{Evaluation of synthetic observations and Novel View Synthesis (NVS).} (a) An end-to-end planner, LTF~\cite{Chitta2023PAMI}, is trained on real data and evaluated on both real and synthetic \texttt{navhard} views, measuring BEV perception (mIoU) and planning (EPDMS) quality. (b) NVS quality is evaluated across several methods using LPIPS on 8 scenes with 10Hz-alternating training and test views.}
    \vspace{-0.5cm}
\end{table}

\boldparagraph{Ablation Study}
Additionally, we evaluate novel view synthesis fidelity using the LPIPS metric~\cite{Zhang2018CVPR} on 8 \texttt{navhard} scenes, where lower scores indicate higher perceptual similarity~\cite{Lindstrom2024CVPR}.
Here, the training and test viewpoints were sampled at alternating 10Hz intervals from the expert trajectory to ensure disjoint inputs and outputs for evaluation.
As shown in Table~\ref{tab:rendering_2}, the baseline Street Gaussians method~\cite{Yan2024ECCV} obtains an LPIPS of 0.354.
Our MTGS-based variant~\cite{Li2025ARXIV} without optimizations improves this score to 0.322, while our full method incorporating LiDAR registration, bundle adjustment, and pose optimization achieves the best LPIPS of 0.253.
Combining these results with the perception and planning evaluations in~\tabref{tab:rendering_1}, we believe our neural rendering pipeline provides sufficient visual fidelity to approximate planning evaluation as in closed-loop settings.

\section{Conclusion}
\label{sec:conclusion}

We introduce pseudo-simulation, a new evaluation paradigm which demonstrates a high correlation to computationally expensive closed-loop simulations.
Our experiments show how it better captures crucial aspects of AV evaluation like error recovery than open-loop evaluation.
Pseudo-simulation offers significant potential impacts for AV development.
It enables more efficient iteration cycles, promotes system robustness by rigorously testing sensitivity to perturbations, and ultimately enhances safety through more comprehensive evaluations.
We hope our public \texttt{navhard} benchmark, featuring pre-rendered data and standardized metrics via an online leaderboard, can foster community adoption of pseudo-simulation for standardized comparisons of AV systems.

\clearpage
\section*{Limitations and Future Work}
While pseudo-simulation demonstrates strong correlation with closed-loop evaluation and offers advantages over existing paradigms, we acknowledge several limitations:

\boldparagraph{Correlation with Real-World Deployment}
Our current validation focuses on establishing correlation with established simulation benchmarks.
We do not yet demonstrate or claim direct correlation with performance metrics from real-world vehicle deployment.
Bridging this gap between simulation-based evaluation and predicting real-world outcomes remains an important direction for future investigation.
Rather than replacing real-world validation, frameworks to augment real-world evaluations with simulation can be applied more effectively with our work~\cite{Luo2025ARXIV}

\boldparagraph{Pre-Processing Computational Cost}
The current pipeline relies on a per-scene optimization process (based on MTGS) to generate the synthetic views, requiring approximately 1-2 hours per scene on current hardware.
While manageable for our dataset scale (under 1000 scenes), this computational cost limits scalability for extremely large datasets.
Exploring recent advancements in potentially faster, feedforward 3D scene representation and rendering methods could offer a path towards significantly reducing this overhead in the future~\cite{Yang2025ICLR,Miao2025CVPR}.

\boldparagraph{Rendering Fidelity and Evaluation}
Despite achieving excellent quantitative results on rendering fidelity (LPIPS) and downstream task performance (mIoU, EPDMS), some visual artifacts may persist in the generated synthetic views.
Our evaluation primarily focuses on algorithmic metrics.
Future work may also benefit from incorporating human perceptual studies to gain a more comprehensive understanding of perceived realism and the potential impact of any remaining artifacts.
Furthermore, combining neural rendering techniques like ours with state-of-the-art generative diffusion models might offer possibilities for enhancing rendering quality~\cite{Yang2024CVPR,Gao2024NEURIPS,Yang2025ARXIV,Wu2025CVPR}.

\boldparagraph{Background Traffic Realism}
The current approach utilizes relatively simple, rule-based traffic models for background agents within the synthetic observations~\cite{Chitta2024ECCV}.
This results in these agents strictly following road-centerline paths during Stage 2 evaluation.
In future work, we aim to incorporate more sophisticated, potentially learned, traffic models that can adapt background agent behavior dynamically based on the ego agent's actions~\cite{Zhang2025CVPR}.
Another possible extension is adversarial background traffic designed to further emphasize the need for robustness~\cite{Hanselmann2022ECCV}.
These extensions could enable the evaluation of more complex, interactive scenarios and improve evaluation fidelity without compromising the scalability of the pseudo-simulation approach.

\boldparagraph{Human Flag Filtering}
Our filtering strategy disregards rule violations also committed by human experts. 
While this helps reduce false positives, it could also risk overlooking important failure and edge cases, since human driving is not always a gold standard for safety. 
Future work could further refine the human flag filtering and explore this trade-off to ensure more reliable evaluation.

\boldparagraph{Metric Design Choices}
We choose multiplicative aggregation because most sub-scores are binary-valued, and multiplication captures compounding failures, e.g., a collision should significantly impact the final score. 
Our Gaussian weighting is selected for its strong empirical performance with minimal assumptions. 
Exploring more principled formulations for aggregation and weighting remains an interesting future direction.

\acknowledgments{
This work was supported by the ERC Starting Grant LEGO-3D (850533), the DFG EXC number 2064/1 - project number 390727645, the German Federal Ministry of Education and Research: Tübingen AI Center, FKZ: 01IS18039A and the German Federal Ministry for Economic Affairs and Energy within the project NXT GEN AI METHODS. We thank the International Max Planck Research School for Intelligent Systems (IMPRS-IS) for supporting Daniel Dauner and Kashyap Chitta. We also thank HuggingFace for hosting our evaluation servers, the team members of OpenDriveLab for their organizational support, Maxim Dolgov for helpful discussions, as well as Napat Karnchanachari and the team from Motional for open-sourcing their dataset and providing us the private test split used in the 2025 NAVSIM Challenges.
}

\bibliography{bibliography_long,bibliography}

\begin{thebibliography}{73}
\providecommand{\natexlab}[1]{#1}
\providecommand{\url}[1]{\texttt{#1}}
\expandafter\ifx\csname urlstyle\endcsname\relax
  \providecommand{\doi}[1]{doi: #1}\else
  \providecommand{\doi}{doi: \begingroup \urlstyle{rm}\Url}\fi

\bibitem[Chen et~al.(2023)Chen, Wu, Chitta, Jaeger, Geiger, and Li]{Chen2024PAMI}
L.~Chen, P.~Wu, K.~Chitta, B.~Jaeger, A.~Geiger, and H.~Li.
\newblock End-to-end autonomous driving: Challenges and frontiers.
\newblock \emph{IEEE Trans. on Pattern Analysis and Machine Intelligence (PAMI)}, 2023.

\bibitem[Buehler et~al.(2009)Buehler, Iagnemma, and Singh]{Buehler2009}
M.~Buehler, K.~Iagnemma, and S.~Singh.
\newblock \emph{The DARPA urban challenge: autonomous vehicles in city traffic}.
\newblock Springer, 2009.

\bibitem[Dosovitskiy et~al.(2017)Dosovitskiy, Ros, Codevilla, Lopez, and Koltun]{Dosovitskiy2017CORL}
A.~Dosovitskiy, G.~Ros, F.~Codevilla, A.~Lopez, and V.~Koltun.
\newblock Carla: An open urban driving simulator.
\newblock In \emph{Proc. Conf. on Robot Learning (CoRL)}, 2017.

\bibitem[{Karnchanachari} et~al.(2024){Karnchanachari}, {Geromichalos}, {Seang Tan}, {Li}, {Eriksen}, {Yaghoubi}, {Mehdipour}, {Bernasconi}, {Kit Fong}, {Guo}, and {Caesar}]{Karnchanachari2024ICRA}
N.~{Karnchanachari}, D.~{Geromichalos}, K.~{Seang Tan}, N.~{Li}, C.~{Eriksen}, S.~{Yaghoubi}, N.~{Mehdipour}, G.~{Bernasconi}, W.~{Kit Fong}, Y.~{Guo}, and H.~{Caesar}.
\newblock {Towards learning-based planning: The nuPlan benchmark for real-world autonomous driving}.
\newblock In \emph{Proc. IEEE International Conf. on Robotics and Automation (ICRA)}, 2024.

\bibitem[Li et~al.(2022)Li, Peng, Feng, Zhang, Xue, and Zhou]{Li2022PAMI}
Q.~Li, Z.~Peng, L.~Feng, Q.~Zhang, Z.~Xue, and B.~Zhou.
\newblock Metadrive: Composing diverse driving scenarios for generalizable reinforcement learning.
\newblock \emph{IEEE Trans. on Pattern Analysis and Machine Intelligence (PAMI)}, 2022.

\bibitem[Weng et~al.(2024)Weng, Ivanovic, Wang, Wang, and Pavone]{Weng2024CVPR}
X.~Weng, B.~Ivanovic, Y.~Wang, Y.~Wang, and M.~Pavone.
\newblock Para-drive: Parallelized architecture for real-time autonomous driving.
\newblock In \emph{Proc. IEEE Conf. on Computer Vision and Pattern Recognition (CVPR)}, 2024.

\bibitem[Li et~al.(2024)Li, Yu, Lan, Li, Kautz, Lu, and Alvarez]{Li2024CVPR}
Z.~Li, Z.~Yu, S.~Lan, J.~Li, J.~Kautz, T.~Lu, and J.~M. Alvarez.
\newblock Is ego status all you need for open-loop end-to-end autonomous driving?
\newblock In \emph{Proc. IEEE Conf. on Computer Vision and Pattern Recognition (CVPR)}, 2024.

\bibitem[Dauner et~al.(2024)Dauner, Hallgarten, Li, Weng, Huang, Yang, Li, Gilitschenski, Ivanovic, Pavone, Geiger, and Chitta]{Dauner2024NEURIPS}
D.~Dauner, M.~Hallgarten, T.~Li, X.~Weng, Z.~Huang, Z.~Yang, H.~Li, I.~Gilitschenski, B.~Ivanovic, M.~Pavone, A.~Geiger, and K.~Chitta.
\newblock Navsim: Data-driven non-reactive autonomous vehicle simulation and benchmarking.
\newblock In \emph{Advances in Neural Information Processing Systems (NeurIPS)}, 2024.

\bibitem[Li et~al.(2025)Li, Qiu, Wu, Lindström, Su, Nießner, and Li]{Li2025ARXIV}
T.~Li, Y.~Qiu, Z.~Wu, C.~Lindström, P.~Su, M.~Nießner, and H.~Li.
\newblock {MTGS}: Multi-traversal gaussian splatting.
\newblock \emph{arXiv.org}, 2503.12552, 2025.

\bibitem[Kerbl et~al.(2023)Kerbl, Kopanas, Leimk{\"u}hler, and Drettakis]{Kerbl2023TOG}
B.~Kerbl, G.~Kopanas, T.~Leimk{\"u}hler, and G.~Drettakis.
\newblock 3d gaussian splatting for real-time radiance field rendering.
\newblock \emph{ACM Transactions on Graphics}, 2023.

\bibitem[Chitta et~al.(2023)Chitta, Prakash, Jaeger, Yu, Renz, and Geiger]{Chitta2023PAMI}
K.~Chitta, A.~Prakash, B.~Jaeger, Z.~Yu, K.~Renz, and A.~Geiger.
\newblock {TransFuser}: Imitation with transformer-based sensor fusion for autonomous driving.
\newblock \emph{IEEE Trans. on Pattern Analysis and Machine Intelligence (PAMI)}, 2023.

\bibitem[Dauner et~al.(2023)Dauner, Hallgarten, Geiger, and Chitta]{Dauner2023CORL}
D.~Dauner, M.~Hallgarten, A.~Geiger, and K.~Chitta.
\newblock Parting with misconceptions about learning-based vehicle motion planning.
\newblock In \emph{Proc. Conf. on Robot Learning (CoRL)}, 2023.

\bibitem[Pitis et~al.(2020)Pitis, Creager, and Garg]{PITIS2020NIPS}
S.~Pitis, E.~Creager, and A.~Garg.
\newblock Counterfactual data augmentation using locally factored dynamics.
\newblock \emph{Advances in Neural Information Processing Systems (NeurIPS)}, 33:\penalty0 3976--3990, 2020.

\bibitem[Pitis et~al.(2022)Pitis, Creager, Mandlekar, and Garg]{PITIS2022NIPS}
S.~Pitis, E.~Creager, A.~Mandlekar, and A.~Garg.
\newblock Mocoda: Model-based counterfactual data augmentation.
\newblock \emph{Advances in Neural Information Processing Systems (NeurIPS)}, 35:\penalty0 18143--18156, 2022.

\bibitem[Chen et~al.(2021)Chen, Xia, and Yu]{CHEN2021ACL}
H.~Chen, R.~Xia, and J.~Yu.
\newblock Reinforced counterfactual data augmentation for dual sentiment classification.
\newblock In \emph{Proceedings of the 2021 Conference on Empirical Methods in Natural Language Processing}, pages 269--278. Association for Computational Linguistics, 2021.

\bibitem[Bojarski et~al.(2016)Bojarski, Del~Testa, Dworakowski, Firner, Flepp, Goyal, Jackel, Monfort, Muller, Zhang, et~al.]{Bojarski2016ARXIV}
M.~Bojarski, D.~Del~Testa, D.~Dworakowski, B.~Firner, B.~Flepp, P.~Goyal, L.~D. Jackel, M.~Monfort, U.~Muller, J.~Zhang, et~al.
\newblock End to end learning for self-driving cars.
\newblock \emph{arXiv.org}, 1604.07316, 2016.

\bibitem[Chen et~al.(2020)Chen, Zhou, Koltun, and Kr{\"a}henb{\"u}hl]{CHEN2020CORL}
D.~Chen, B.~Zhou, V.~Koltun, and P.~Kr{\"a}henb{\"u}hl.
\newblock Learning by cheating.
\newblock In \emph{Proc. Conf. on Robot Learning (CoRL)}, 2020.

\bibitem[Codevilla et~al.(2019)Codevilla, Santana, L{\'o}pez, and Gaidon]{Codevilla2019ICCV}
F.~Codevilla, E.~Santana, A.~M. L{\'o}pez, and A.~Gaidon.
\newblock Exploring the limitations of behavior cloning for autonomous driving.
\newblock In \emph{Proc. of the IEEE International Conf. on Computer Vision (ICCV)}, 2019.

\bibitem[Prakash et~al.(2020)Prakash, Behl, Ohn-Bar, Chitta, and Geiger]{Prakash2020CVPR}
A.~Prakash, A.~Behl, E.~Ohn-Bar, K.~Chitta, and A.~Geiger.
\newblock Exploring data aggregation in policy learning for vision-based urban autonomous driving.
\newblock In \emph{Proc. IEEE Conf. on Computer Vision and Pattern Recognition (CVPR)}, 2020.

\bibitem[Jaeger et~al.(2023)Jaeger, Chitta, and Geiger]{Jaeger2023ICCV}
B.~Jaeger, K.~Chitta, and A.~Geiger.
\newblock Hidden biases of end-to-end driving models.
\newblock In \emph{Proc. of the IEEE International Conf. on Computer Vision (ICCV)}, 2023.

\bibitem[Ma et~al.(2024)Ma, Zhou, Tang, Zhang, Han, Jiang, Zhan, Jia, Lang, Sun, et~al.]{MA2024ARXIV}
E.~Ma, L.~Zhou, T.~Tang, Z.~Zhang, D.~Han, J.~Jiang, K.~Zhan, P.~Jia, X.~Lang, H.~Sun, et~al.
\newblock Unleashing generalization of end-to-end autonomous driving with controllable long video generation.
\newblock \emph{arXiv.org}, 2406.01349, 2024.

\bibitem[Zimmerlin et~al.(2024)Zimmerlin, Bei{\ss}wenger, Jaeger, Geiger, and Chitta]{ZIMMERLIN2024ARXIV}
J.~Zimmerlin, J.~Bei{\ss}wenger, B.~Jaeger, A.~Geiger, and K.~Chitta.
\newblock Hidden biases of end-to-end driving datasets.
\newblock \emph{arXiv.org}, 2412.09602, 2024.

\bibitem[Renz et~al.(2025)Renz, Chen, Arani, and Sinavski]{RENZ2025ARXIV}
K.~Renz, L.~Chen, E.~Arani, and O.~Sinavski.
\newblock Simlingo: Vision-only closed-loop autonomous driving with language-action alignment.
\newblock \emph{arXiv.org}, 2503.09594, 2025.

\bibitem[Gulino et~al.(2023)Gulino, Fu, Luo, Tucker, Bronstein, Lu, Harb, Pan, Wang, Chen, Co-Reyes, Agarwal, Roelofs, Lu, Montali, Mougin, Yang, White, Faust, McAllister, Anguelov, and Sapp]{Gulino2023NIPS}
C.~Gulino, J.~Fu, W.~Luo, G.~Tucker, E.~Bronstein, Y.~Lu, J.~Harb, X.~Pan, Y.~Wang, X.~Chen, J.~D. Co-Reyes, R.~Agarwal, R.~Roelofs, Y.~Lu, N.~Montali, P.~Mougin, Z.~Yang, B.~White, A.~Faust, R.~McAllister, D.~Anguelov, and B.~Sapp.
\newblock Waymax: An accelerated, data-driven simulator for large-scale autonomous driving research.
\newblock In \emph{Advances in Neural Information Processing Systems (NeurIPS)}, 2023.

\bibitem[Vinitsky et~al.(2022)Vinitsky, Lichtl{\'e}, Yang, Amos, and Foerster]{Vinitsky2022NIPS}
E.~Vinitsky, N.~Lichtl{\'e}, X.~Yang, B.~Amos, and J.~Foerster.
\newblock Nocturne: a scalable driving benchmark for bringing multi-agent learning one step closer to the real world.
\newblock In \emph{Advances in Neural Information Processing Systems (NeurIPS)}, 2022.

\bibitem[Kazemkhani et~al.(2024)Kazemkhani, Pandya, Cornelisse, Shacklett, and Vinitsky]{Kazemkhani2024ARXIV}
S.~Kazemkhani, A.~Pandya, D.~Cornelisse, B.~Shacklett, and E.~Vinitsky.
\newblock Gpudrive: Data-driven, multi-agent driving simulation at 1 million fps.
\newblock \emph{arXiv.org}, 2408.01584, 2024.

\bibitem[Amini et~al.(2020)Amini, Gilitschenski, Phillips, Moseyko, Banerjee, Karaman, and Rus]{Amini2020RAL}
A.~Amini, I.~Gilitschenski, J.~Phillips, J.~Moseyko, R.~Banerjee, S.~Karaman, and D.~Rus.
\newblock Learning robust control policies for end-to-end autonomous driving from data-driven simulation.
\newblock \emph{IEEE Robotics and Automation Letters (RA-L)}, 2020.

\bibitem[Amini et~al.(2022)Amini, Wang, Gilitschenski, Schwarting, Liu, Han, Karaman, and Rus]{Amini2022ICRA}
A.~Amini, T.-H. Wang, I.~Gilitschenski, W.~Schwarting, Z.~Liu, S.~Han, S.~Karaman, and D.~Rus.
\newblock Vista 2.0: An open, data-driven simulator for multimodal sensing and policy learning for autonomous vehicles.
\newblock In \emph{Proc. IEEE International Conf. on Robotics and Automation (ICRA)}, 2022.

\bibitem[Wang et~al.(2022)Wang, Amini, Schwarting, Gilitschenski, Karaman, and Rus]{Wang2022ICRA}
T.-H. Wang, A.~Amini, W.~Schwarting, I.~Gilitschenski, S.~Karaman, and D.~Rus.
\newblock Learning interactive driving policies via data-driven simulation.
\newblock In \emph{Proc. IEEE International Conf. on Robotics and Automation (ICRA)}, 2022.

\bibitem[Yang et~al.(2023)Yang, Chen, Wang, Manivasagam, Ma, Yang, and Urtasun]{Yang2023CVPR}
Z.~Yang, Y.~Chen, J.~Wang, S.~Manivasagam, W.-C. Ma, A.~J. Yang, and R.~Urtasun.
\newblock Unisim: A neural closed-loop sensor simulator.
\newblock In \emph{Proc. IEEE Conf. on Computer Vision and Pattern Recognition (CVPR)}, 2023.

\bibitem[Tonderski et~al.(2024)Tonderski, Lindstr{\"o}m, Hess, Ljungbergh, Svensson, and Petersson]{Tonderski2024CVPR}
A.~Tonderski, C.~Lindstr{\"o}m, G.~Hess, W.~Ljungbergh, L.~Svensson, and C.~Petersson.
\newblock {NeuRAD}: Neural rendering for autonomous driving.
\newblock In \emph{Proc. IEEE Conf. on Computer Vision and Pattern Recognition (CVPR)}, 2024.

\bibitem[Ljungbergh et~al.(2024)Ljungbergh, Tonderski, Johnander, Caesar, {\AA}str{\"o}m, Felsberg, and Petersson]{Ljungbergh2024ECCV}
W.~Ljungbergh, A.~Tonderski, J.~Johnander, H.~Caesar, K.~{\AA}str{\"o}m, M.~Felsberg, and C.~Petersson.
\newblock Neuroncap: Photorealistic closed-loop safety testing for autonomous driving.
\newblock In \emph{Proc. of the European Conf. on Computer Vision (ECCV)}, 2024.

\bibitem[Yang et~al.(2024)Yang, Wen, Ma, Mei, Li, Wei, Lei, Fu, Cai, Dou, et~al.]{Yang2024ARXIV}
X.~Yang, L.~Wen, Y.~Ma, J.~Mei, X.~Li, T.~Wei, W.~Lei, D.~Fu, P.~Cai, M.~Dou, et~al.
\newblock Drivearena: A closed-loop generative simulation platform for autonomous driving.
\newblock \emph{arXiv.org}, 2408.00415, 2024.

\bibitem[Zhou et~al.(2024)Zhou, Lin, Wang, Lu, Bai, Liu, Wang, Geiger, and Liao]{Zhou2024ARXIV}
H.~Zhou, L.~Lin, J.~Wang, Y.~Lu, D.~Bai, B.~Liu, Y.~Wang, A.~Geiger, and Y.~Liao.
\newblock Hugsim: A real-time, photo-realistic and closed-loop simulator for autonomous driving.
\newblock \emph{arXiv.org}, 2412.01718, 2024.

\bibitem[You et~al.(2024)You, Jia, Zhang, Zhu, and Yan]{You2024ARXIV}
J.~You, X.~Jia, Z.~Zhang, Y.~Zhu, and J.~Yan.
\newblock Bench2drive-r: Turning real world data into reactive closed-loop autonomous driving benchmark by generative model.
\newblock \emph{arXiv.org}, 2412.09647, 2024.

\bibitem[Sima et~al.(2024)Sima, Renz, Chitta, Chen, Zhang, Xie, Beißwenger, Luo, Geiger, and Li]{Sima2024ECCV}
C.~Sima, K.~Renz, K.~Chitta, L.~Chen, H.~Zhang, C.~Xie, J.~Beißwenger, P.~Luo, A.~Geiger, and H.~Li.
\newblock Drivelm: Driving with graph visual question answering.
\newblock In \emph{Proc. of the European Conf. on Computer Vision (ECCV)}, 2024.

\bibitem[Codevilla et~al.(2018)Codevilla, Lopez, Koltun, and Dosovitskiy]{Codevilla2018ECCV}
F.~Codevilla, A.~M. Lopez, V.~Koltun, and A.~Dosovitskiy.
\newblock On offline evaluation of vision-based driving models.
\newblock In \emph{Proc. of the European Conf. on Computer Vision (ECCV)}, 2018.

\bibitem[Zhai et~al.(2023)Zhai, Feng, Du, Mao, Liu, Tan, Zhang, Ye, and Wang]{Zhai2023ARXIV}
J.-T. Zhai, Z.~Feng, J.~Du, Y.~Mao, J.-J. Liu, Z.~Tan, Y.~Zhang, X.~Ye, and J.~Wang.
\newblock Rethinking the open-loop evaluation of end-to-end autonomous driving in nuscenes.
\newblock \emph{arXiv.org}, 2305.10430, 2023.

\bibitem[Caesar et~al.(2020)Caesar, Bankiti, Lang, Vora, Liong, Xu, Krishnan, Pan, Baldan, and Beijbom]{Caesar2021CVPR}
H.~Caesar, V.~Bankiti, A.~H. Lang, S.~Vora, V.~E. Liong, Q.~Xu, A.~Krishnan, Y.~Pan, G.~Baldan, and O.~Beijbom.
\newblock nuscenes: A multimodal dataset for autonomous driving.
\newblock In \emph{Proc. IEEE Conf. on Computer Vision and Pattern Recognition (CVPR)}, 2020.

\bibitem[Wen et~al.(2020)Wen, Lin, Darrell, Jayaraman, and Gao]{Wen2020NEURIPS}
C.~Wen, J.~Lin, T.~Darrell, D.~Jayaraman, and Y.~Gao.
\newblock Fighting copycat agents in behavioral cloning from observation histories.
\newblock In \emph{Advances in Neural Information Processing Systems (NeurIPS)}, 2020.

\bibitem[Rajamani(2011)]{Rajamani2011Springer}
R.~Rajamani.
\newblock \emph{Vehicle dynamics and control}.
\newblock Springer Science \& Business Media, 2011.

\bibitem[Lehtomaki et~al.(1981)Lehtomaki, Sandell, and Athans]{Lehtomaki1981TAC}
N.~Lehtomaki, N.~Sandell, and M.~Athans.
\newblock Robustness results in linear-quadratic gaussian based multivariable control designs.
\newblock \emph{IEEE Trans. on Automatic Control (TAC)}, 1981.

\bibitem[Treiber et~al.(2000)Treiber, Hennecke, and Helbing]{Treiber2000}
M.~Treiber, A.~Hennecke, and D.~Helbing.
\newblock Congested traffic states in empirical observations and microscopic simulations.
\newblock \emph{Physical review E}, 2000.

\bibitem[{Li} et~al.(2025){Li}, {Li}, {Lan}, {Xie}, {Zhang}, {Liu}, {Wu}, {Yu}, and {Alvarez}]{Li2025ARXIVb}
K.~{Li}, Z.~{Li}, S.~{Lan}, Y.~{Xie}, Z.~{Zhang}, J.~{Liu}, Z.~{Wu}, Z.~{Yu}, and J.~M. {Alvarez}.
\newblock {Hydra-MDP++: Advancing End-to-End Driving via Expert-Guided Hydra-Distillation}.
\newblock \emph{arXiv.org}, 2503.12820, 2025.

\bibitem[Ost et~al.(2021)Ost, Mannan, Thuerey, Knodt, and Heide]{Ost2021CVPR}
J.~Ost, F.~Mannan, N.~Thuerey, J.~Knodt, and F.~Heide.
\newblock Neural scene graphs for dynamic scenes.
\newblock In \emph{Proc. IEEE Conf. on Computer Vision and Pattern Recognition (CVPR)}, 2021.

\bibitem[Vizzo et~al.(2023)Vizzo, Guadagnino, Mersch, Wiesmann, Behley, and Stachniss]{Vizzo2023RAL}
I.~Vizzo, T.~Guadagnino, B.~Mersch, L.~Wiesmann, J.~Behley, and C.~Stachniss.
\newblock {KISS-ICP: In Defense of Point-to-Point ICP -- Simple, Accurate, and Robust Registration If Done the Right Way}.
\newblock \emph{RA-L}, 2023.

\bibitem[Sch\"{o}nberger and Frahm(2016)]{Schoenberger2016CVPR}
J.~L. Sch\"{o}nberger and J.-M. Frahm.
\newblock Structure-from-motion revisited.
\newblock In \emph{Proc. IEEE Conf. on Computer Vision and Pattern Recognition (CVPR)}, 2016.

\bibitem[Yan et~al.(2024)Yan, Qu, Xu, Zhao, Wang, Wang, and Li]{Yan2024CVPR}
C.~Yan, D.~Qu, D.~Xu, B.~Zhao, Z.~Wang, D.~Wang, and X.~Li.
\newblock {GS-SLAM}: Dense visual slam with 3d gaussian splatting.
\newblock In \emph{Proc. IEEE Conf. on Computer Vision and Pattern Recognition (CVPR)}, 2024.

\bibitem[Renz et~al.(2022)Renz, Chitta, Mercea, Koepke, Akata, and Geiger]{Renz2022CORL}
K.~Renz, K.~Chitta, O.-B. Mercea, S.~Koepke, Z.~Akata, and A.~Geiger.
\newblock Plant: Explainable planning transformers via object-level representations.
\newblock In \emph{Proc. Conf. on Robot Learning (CoRL)}, 2022.

\bibitem[Scheel et~al.(2021)Scheel, Bergamini, Wolczyk, Osi{\'n}ski, and Ondruska]{Scheel2021CORL}
O.~Scheel, L.~Bergamini, M.~Wolczyk, B.~Osi{\'n}ski, and P.~Ondruska.
\newblock Urban driver: Learning to drive from real-world demonstrations using policy gradients.
\newblock In \emph{Proc. Conf. on Robot Learning (CoRL)}, 2021.

\bibitem[Nguyen et~al.(2026)Nguyen, Fauth, Jaeger, Dauner, Igl, Geiger, and Chitta]{Nguyen2026CVPR}
L.~Nguyen, M.~Fauth, B.~Jaeger, D.~Dauner, M.~Igl, A.~Geiger, and K.~Chitta.
\newblock Lead: Minimizing learner-expert asymmetry in end-to-end driving.
\newblock In \emph{Proc. IEEE Conf. on Computer Vision and Pattern Recognition (CVPR)}, 2026.

\bibitem[Feng et~al.(2025)Feng, Gao, Zablocki, Li, Li, Liu, Cord, and Alahi]{Feng2025ARXIV}
L.~Feng, Y.~Gao, E.~Zablocki, Q.~Li, W.~Li, S.~Liu, M.~Cord, and A.~Alahi.
\newblock Rap: 3d rasterization augmented end-to-end planning.
\newblock \emph{arXiv.org}, 2510.04333, 2025.

\bibitem[Li et~al.(2025)Li, Yao, Wang, Sun, Chen, Chang, Shen, Song, Wu, Lan, and Alvarez]{Li2025ARXIVc}
Z.~Li, W.~Yao, Z.~Wang, X.~Sun, J.~Chen, N.~Chang, M.~Shen, J.~Song, Z.~Wu, S.~Lan, and J.~M. Alvarez.
\newblock Ztrs: Zero-imitation end-to-end autonomous driving with trajectory scoring.
\newblock \emph{arXiv.org}, 2510.24108, 2025.

\bibitem[Liu et~al.(2025)Liu, Jia, Yu, Song, Li, Jia, Wu, Hao, and Luo]{Liu2025ARXIV}
L.~Liu, C.~Jia, G.~Yu, Z.~Song, J.~Li, F.~Jia, P.~Wu, X.~Hao, and Y.~Luo.
\newblock Guideflow: Constraint-guided flow matching for planning in end-to-end autonomous driving.
\newblock \emph{arXiv.org}, 2511.18729, 2025.

\bibitem[Tian et~al.(2026)Tian, Li, Liu, Yang, Qiu, Li, Wang, Gao, Zhang, Wang, Ye, Tan, Chen, and Li]{Tian2026CVPR}
H.~Tian, T.~Li, H.~Liu, J.~Yang, Y.~Qiu, G.~Li, J.~Wang, Y.~Gao, Z.~Zhang, L.~Wang, H.~Ye, T.~Tan, L.~Chen, and H.~Li.
\newblock Simscale: Learning to drive via real-world simulation at scale.
\newblock In \emph{Proc. IEEE Conf. on Computer Vision and Pattern Recognition (CVPR)}, 2026.

\bibitem[Kirby et~al.(2026)Kirby, Boulch, Xu, Yin, Puy, Zablocki, Bursuc, Gidaris, Marlet, Bartoccioni, Cao, Samet, VU, and Cord]{Kirby2026CVPR}
E.~Kirby, A.~Boulch, Y.~Xu, Y.~Yin, G.~Puy, {\'E}.~Zablocki, A.~Bursuc, S.~Gidaris, R.~Marlet, F.~Bartoccioni, A.-Q. Cao, N.~Samet, T.-H. VU, and M.~Cord.
\newblock Driving on registers.
\newblock In \emph{Proc. IEEE Conf. on Computer Vision and Pattern Recognition (CVPR)}, 2026.

\bibitem[Li et~al.(2022)Li, Wang, Li, Xie, Sima, Lu, Qiao, and Dai]{Li2022ECCV}
Z.~Li, W.~Wang, H.~Li, E.~Xie, C.~Sima, T.~Lu, Y.~Qiao, and J.~Dai.
\newblock {BEVFormer}: Learning bird’s-eye-view representation from multi-camera images via spatiotemporal transformers.
\newblock In \emph{Proc. of the European Conf. on Computer Vision (ECCV)}, 2022.

\bibitem[Li et~al.(2024)Li, Li, Wang, Lan, Yu, Ji, Li, Zhu, Kautz, Wu, Jiang, and Alvarez]{Li2024ARXIV}
Z.~Li, K.~Li, S.~Wang, S.~Lan, Z.~Yu, Y.~Ji, Z.~Li, Z.~Zhu, J.~Kautz, Z.~Wu, Y.-G. Jiang, and J.~M. Alvarez.
\newblock Hydra-mdp: End-to-end multimodal planning with multi-target hydra-distillation.
\newblock \emph{arXiv.org}, 2024.

\bibitem[Behl et~al.(2020)Behl, Chitta, Prakash, Ohn-Bar, and Geiger]{Behl2020IROS}
A.~Behl, K.~Chitta, A.~Prakash, E.~Ohn-Bar, and A.~Geiger.
\newblock Label efficient visual abstractions for autonomous driving.
\newblock In \emph{Proc. IEEE International Conf. on Intelligent Robots and Systems (IROS)}, 2020.

\bibitem[Schreier et~al.(2023)Schreier, Renz, Geiger, and Chitta]{Schreier2023ICCVW}
T.~Schreier, K.~Renz, A.~Geiger, and K.~Chitta.
\newblock On offline evaluation of 3d object detection for autonomous driving.
\newblock In \emph{Proc. of the IEEE International Conf. on Computer Vision (ICCV) Workshops}, 2023.

\bibitem[Yan et~al.(2024)Yan, Lin, Zhou, Wang, Sun, Zhan, Lang, Zhou, and Peng]{Yan2024ECCV}
Y.~Yan, H.~Lin, C.~Zhou, W.~Wang, H.~Sun, K.~Zhan, X.~Lang, X.~Zhou, and S.~Peng.
\newblock {Street Gaussians}: Modeling dynamic urban scenes with gaussian splatting.
\newblock In \emph{Proc. of the European Conf. on Computer Vision (ECCV)}, 2024.

\bibitem[Zhang et~al.(2018)Zhang, Isola, Efros, Shechtman, and Wang]{Zhang2018CVPR}
R.~Zhang, P.~Isola, A.~A. Efros, E.~Shechtman, and O.~Wang.
\newblock The unreasonable effectiveness of deep features as a perceptual metric.
\newblock In \emph{Proc. IEEE Conf. on Computer Vision and Pattern Recognition (CVPR)}, 2018.

\bibitem[Lindstr{\"o}m et~al.(2024)Lindstr{\"o}m, Hess, Lilja, Fatemi, Hammarstrand, Petersson, and Svensson]{Lindstrom2024CVPR}
C.~Lindstr{\"o}m, G.~Hess, A.~Lilja, M.~Fatemi, L.~Hammarstrand, C.~Petersson, and L.~Svensson.
\newblock Are nerfs ready for autonomous driving? towards closing the real-to-simulation gap.
\newblock In \emph{Proc. IEEE Conf. on Computer Vision and Pattern Recognition (CVPR)}, 2024.

\bibitem[Luo et~al.(2025)Luo, Yang, Watson, Sharma, Veer, Schmerling, and Pavone]{Luo2025ARXIV}
R.~Luo, H.~Yang, M.~Watson, A.~Sharma, S.~Veer, E.~Schmerling, and M.~Pavone.
\newblock Leveraging correlation across test platforms for variance-reduced metric estimation.
\newblock \emph{arXiv.org}, 2506.20553, 2025.

\bibitem[Yang et~al.(2025)Yang, Huang, Chen, Wang, Li, You, Igl, Sharma, Karkus, Xu, Ivanovic, Wang, and Pavone]{Yang2025ICLR}
J.~Yang, J.~Huang, Y.~Chen, Y.~Wang, B.~Li, Y.~You, M.~Igl, A.~Sharma, P.~Karkus, D.~Xu, B.~Ivanovic, Y.~Wang, and M.~Pavone.
\newblock Storm: Spatio-temporal reconstruction model for large-scale outdoor scenes.
\newblock In \emph{Proc. of the International Conf. on Learning Representations (ICLR)}, 2025.

\bibitem[Miao et~al.(2025)Miao, Huang, Bai, Yan, Zhou, Wang, Liu, Geiger, and Liao]{Miao2025CVPR}
S.~Miao, J.~Huang, D.~Bai, X.~Yan, H.~Zhou, Y.~Wang, B.~Liu, A.~Geiger, and Y.~Liao.
\newblock Evolsplat: Efficient volume-based gaussian splatting for urban view synthesis.
\newblock In \emph{Proc. IEEE Conf. on Computer Vision and Pattern Recognition (CVPR)}, 2025.

\bibitem[{Yang} et~al.(2024){Yang}, {Gao}, {Qiu}, {Chen}, {Li}, {Dai}, {Chitta}, {Wu}, {Zeng}, {Luo}, {Zhang}, {Geiger}, {Qiao}, and {Li}]{Yang2024CVPR}
J.~{Yang}, S.~{Gao}, Y.~{Qiu}, L.~{Chen}, T.~{Li}, B.~{Dai}, K.~{Chitta}, P.~{Wu}, J.~{Zeng}, P.~{Luo}, J.~{Zhang}, A.~{Geiger}, Y.~{Qiao}, and H.~{Li}.
\newblock {Generalized Predictive Model for Autonomous Driving}.
\newblock In \emph{Proc. IEEE Conf. on Computer Vision and Pattern Recognition (CVPR)}, 2024.

\bibitem[Gao et~al.(2024)Gao, Yang, Chen, Chitta, Qiu, Geiger, Zhang, and Li]{Gao2024NEURIPS}
S.~Gao, J.~Yang, L.~Chen, K.~Chitta, Y.~Qiu, A.~Geiger, J.~Zhang, and H.~Li.
\newblock Vista: A generalizable driving world model with high fidelity and versatile controllability.
\newblock In \emph{Advances in Neural Information Processing Systems (NeurIPS)}, 2024.

\bibitem[{Yang} et~al.(2025){Yang}, {Chitta}, {Gao}, {Chen}, {Shao}, {Jia}, {Li}, {Geiger}, {Yue}, and {Chen}]{Yang2025ARXIV}
J.~{Yang}, K.~{Chitta}, S.~{Gao}, L.~{Chen}, Y.~{Shao}, X.~{Jia}, H.~{Li}, A.~{Geiger}, X.~{Yue}, and L.~{Chen}.
\newblock Resim: Reliable world simulation for autonomous drivingend to end learning for self-driving cars.
\newblock \emph{arXiv.org}, 2506.09981, 2025.

\bibitem[Wu et~al.(2025)Wu, Zhang, Turki, Ren, Gao, Shou, Fidler, Gojcic, and Ling]{Wu2025CVPR}
J.~Z. Wu, Y.~Zhang, H.~Turki, X.~Ren, J.~Gao, M.~Z. Shou, S.~Fidler, Z.~Gojcic, and H.~Ling.
\newblock Difix3d+: Improving 3d reconstructions with single-step diffusion models.
\newblock In \emph{Proc. IEEE Conf. on Computer Vision and Pattern Recognition (CVPR)}, 2025.

\bibitem[Chitta et~al.(2024)Chitta, Dauner, and Geiger]{Chitta2024ECCV}
K.~Chitta, D.~Dauner, and A.~Geiger.
\newblock Sledge: Synthesizing driving environments with generative models and rule-based traffic.
\newblock In \emph{Proc. of the European Conf. on Computer Vision (ECCV)}, 2024.

\bibitem[Zhang et~al.(2025)Zhang, Karkus, Igl, Ding, Chen, Ivanovic, and Pavone]{Zhang2025CVPR}
Z.~Zhang, P.~Karkus, M.~Igl, W.~Ding, Y.~Chen, B.~Ivanovic, and M.~Pavone.
\newblock Closed-loop supervised fine-tuning of tokenized traffic models.
\newblock In \emph{Proc. IEEE Conf. on Computer Vision and Pattern Recognition (CVPR)}, 2025.

\bibitem[Hanselmann et~al.(2022)Hanselmann, Renz, Chitta, Bhattacharyya, and Geiger]{Hanselmann2022ECCV}
N.~Hanselmann, K.~Renz, K.~Chitta, A.~Bhattacharyya, and A.~Geiger.
\newblock King: Generating safety-critical driving scenarios for robust imitation via kinematics gradients.
\newblock In \emph{Proc. of the European Conf. on Computer Vision (ECCV)}, 2022.

\end{thebibliography}

\end{document}